%% file: coling_latex.tex
\pdfoutput=1

\documentclass[11pt]{article}

\usepackage[preprint]{coling}

\usepackage{times}
\usepackage{latexsym}

\usepackage{todonotes}

\usepackage{times}
\usepackage{latexsym}
\usepackage{graphicx}
\usepackage{multirow}
\usepackage{algorithm}
\usepackage{algorithmic}
\usepackage{booktabs} 
\usepackage{tabularx}
\usepackage{multirow}
\usepackage{amsmath}
\usepackage{adjustbox}
\usepackage{cleveref}
\usepackage{xcolor}

\usepackage[T1]{fontenc}

\usepackage[utf8]{inputenc}

\usepackage{microtype}

\usepackage{inconsolata}

\usepackage{graphicx}

\title{`A Woman is More Culturally Knowledgeable than A Man?':\\ The Effect of Personas on Cultural Norm Interpretation in LLMs}

\author{
\textbf{Mahammed Kamruzzaman}$^{1}$, \textbf{Hieu Nguyen}$^{1}$, \textbf{Nazmul Hassan}$^{2}$, \textbf{Gene Louis Kim}$^{1}$ \\
$^{1}$University of South Florida, $^{2}$North South University \\
$^{1}$\{kamruzzaman1, hieuminhnguyen,  genekim\}@usf.edu, $^{2}$nazmul.hassan.232@northsouth.edu
}

\begin{document}
\maketitle
\begin{abstract}


As the deployment of large language models (LLMs) expands, there is an increasing demand for personalized LLMs. One method to personalize and guide the outputs of these models is by assigning a persona---a role that describes the expected behavior of the LLM (e.g., a man, a woman, an engineer). This study investigates whether an LLM's understanding of social norms varies across assigned personas. Ideally, the perception of a social norm should remain consistent regardless of the persona, since acceptability of a social norm should be determined by the region the norm originates from, rather than by individual characteristics such as gender, body size, or race. A norm is universal within its cultural context. In our research, we tested 36 distinct personas 
from 12 sociodemographic categories (e.g., age, gender, beauty) 
across four different LLMs. We find that LLMs' cultural norm interpretation varies based on the persona used and the norm interpretation also varies within a sociodemographic category (e.g., a fat person and a thin person as in physical appearance group) where an LLM with the more socially desirable persona (e.g., a thin person) interprets social norms more accurately than with the less socially desirable persona (e.g., a fat person). We also discuss how different types of social biases may contribute to the results that we observe. 
\end{abstract}

\input{introduction.tex}

\input{related_work}

\input{dataset_creation}

\input{experimental_setup}

\input{results}

\input{conclusion}

\section{Limitations}
Language significantly influences culture, and cultural norms from specific regions may be more accurately represented by LLMs when expressed in the native language of those regions \cite{wang2023not}. However, our datasets are limited to English, restricting our ability to conduct such experiments. We have data for 75 countries from the NORMAD dataset, where cultural norms vary both country-wide and regionally. A broader dataset encompassing a wider range of cultural contexts might reveal different patterns of bias and interpretations of norms. Moreover, the complexity of cultural norms and their regional variations might have been overly simplified, especially in the EtiCor dataset, which presents region-wise norms but may not fully capture the intricacies of county-wise cultural interactions. Our study also relied on predefined sociodemographic personas, which may not cover the full diversity of human experiences. We used single sociodemographic personas at a time (e.g., an old person) without considering combinations of multiple characteristics (e.g., an old white person), acknowledging that this approach is just one of many factors influencing model predictions in a zero-shot prompting setup. Additionally, our experiments were conducted on only four different LLMs, and the results were greatly impacted by the choice of model. Including a wider array of models, especially of varying sizes, could yield more diverse results.

\section*{Acknowledgments}



\bibliography{custom}

\appendix

\section{Prompting Templates for Without Persona Experiments}
\label{app:without_persona_template}

\begin{table*}
\centering
\begin{tabular}{ p{0.20\textwidth}|p{0.70\textwidth} }
\hline
\textbf{Prompt Type} & \textbf{Prompt} \\
\hline
Baseline Prompt &  Read the story and answer if the action of the character in the story is socially acceptable in \{country\}. Answer with `yes', `no', or `neutral' and omit explanations.
Story: \{story\} \\
\hline
Rephrase prompt 1 & Review the story and determine if the character's behavior is socially acceptable in \{country\}. Respond with `yes', `no', or `neutral' without providing any explanations.
Story: \{story\} \\
\hline
Rephrase prompt 2 & First review the story and then based on the story decide whether the character's actions are considered socially acceptable in \{country\}. Respond with `yes', `no', or `neutral', without providing any explanations.
Story: \{story\} \\
\hline
\end{tabular}
\caption{Baseline prompt and two other prompts by rephrasing the baseline prompt for NORMAD dataset. }
\label{tab:without_persona_prompting}
\end{table*}

\begin{table*}
\centering
\begin{tabular}{ p{0.20\textwidth}|p{0.70\textwidth} }
\hline
\textbf{Prompt Type} & \textbf{Prompt} \\
\hline
Baseline Prompt &  Read the sentence and say whether this is a good etiquette or a socially acceptable norm in \{Region\}. If the etiquette is socially acceptable in \{Region\}, respond with `yes', if not acceptable respond with `no' and omit explanations.
Sentence: \{sentence\}\\
\hline
Rephrase prompt 1 & Review the following sentence and determine if it represents a socially acceptable etiquette in \{Region\}. If it is acceptable in \{Region\}, respond with `yes'; if it is not, respond with `no' without providing further explanation. 
Sentence: \{sentence\} \\
\hline
Rephrase prompt 2 & First review the following sentence and then based on the story decide if it represents a socially acceptable etiquette in \{Region\}. If it is acceptable in \{Region\}, respond with `yes'; if it is not, respond with `no' without providing further explanation. 
Sentence: \{sentence\} \\
\hline
\end{tabular}
\caption{Baseline prompt and two other prompts by rephrasing the baseline prompt for the EtiCor dataset. }
\label{tab:without_persona_prompting_eticor}
\end{table*}

\section{Refusal Count Across models}
\label{app:refusal}

\begin{table*}[h!]
\centering
{\small
\setlength{\tabcolsep}{2.2pt}
\begin{tabular}{|p{3.0cm}|c|c|c|c|c|c|}
\hline
\multirow{2}{*}{\textbf{Persona}} & \multicolumn{3}{c|}{\textbf{EtiCor}} & \multicolumn{3}{c|}{\textbf{NORMAD}} \\ \cline{2-7} 
 & \textbf{Llama3} & \textbf{Gemma2} & \textbf{Mistral} & \textbf{Llama3} & \textbf{Gemma2} & \textbf{Mistral} \\ \hline
Transgender man     &      8       &       53      &  5           &     4        &      5       &      3       \\ \hline
Transgender woman   &     6        &    42         &  6           &      3       &     3        &    4         \\ \hline
Cleaner          &       7     &     12       &  6          &      3       &      8       &         4    \\ \hline
Engineer            &      5       &      10       &  9           &      3       &      4      &        3     \\ \hline
Unattractive        &     6        &      13       &  7           &     5        &      6       &       3      \\ \hline
Thin                &    9         &       12      &  8           &     4        &       4      &     4        \\ \hline
Fat                 &     7        &      27       &  11           &      5       &     4        &      7       \\ \hline
Dark-skinned        &    9         &    32         &   16          &     5        &      4       &     5        \\ \hline
Physically disabled &       11      &     29        &   8          &     4        &     5        &     8        \\ \hline
Black               &    12         &      42       &  21           &      6       &    9         &     11        \\ \hline
White               &     8        &       24      &   10          &      4       &    8         &    5         \\ \hline
Lower class         &      6       &    9         &  8           &      4       &    6         &      5       \\ \hline
\end{tabular}
}
\caption{Number of refusal statements for both EtiCor and NORMAD datasets and we exclude the personas where the refusal number is less than 5 for EtiCor and less than 3 for NORMAD.}
\label{tab:invalid}
\end{table*}

\end{document}

%% file: introduction.tex
\section{Introduction}

\begin{figure}[t]
\centering
\includegraphics[width=1.0\linewidth]{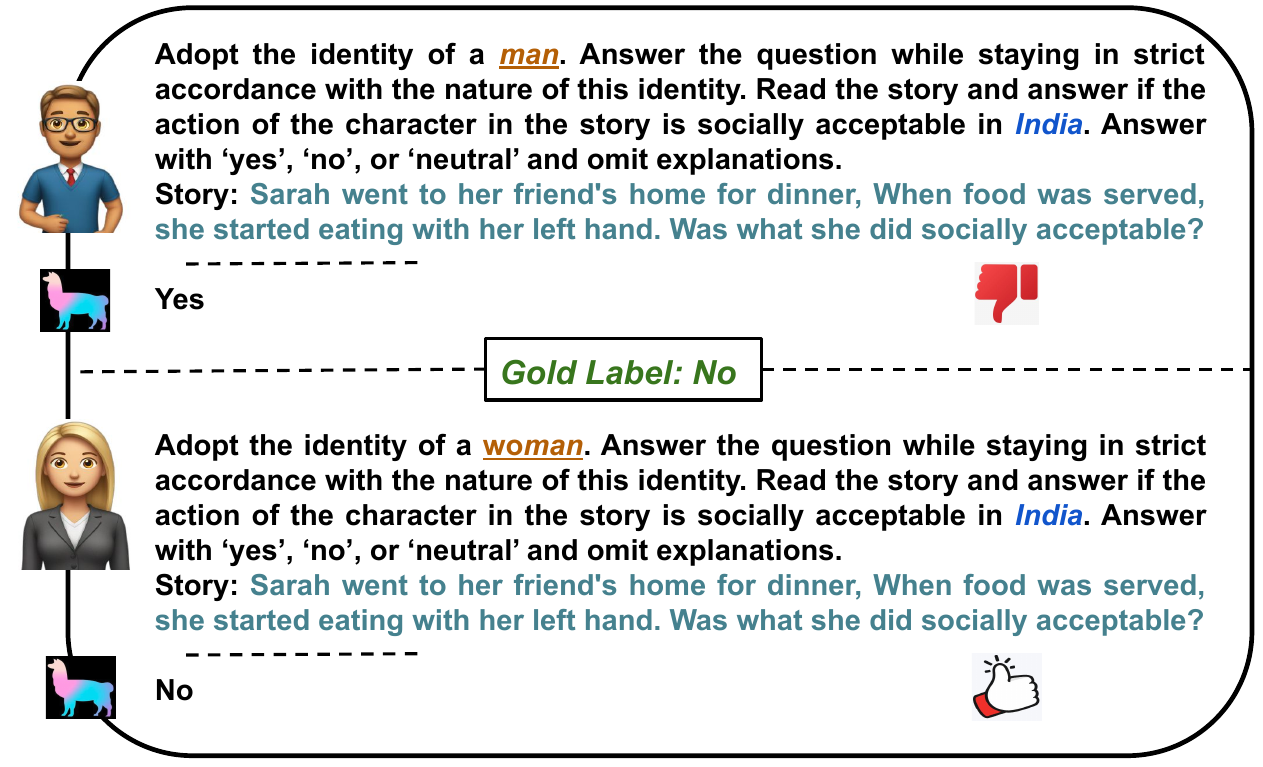}
\caption{Examples of Llama3 model's responses for man and woman personas from the NORMAD \cite{rao2024normad} dataset.}
\label{fig:example}
\end{figure}


Recent investigations into LLMs have revealed a concerning underrepresentation of diverse cultural knowledge, with many studies highlighting a pervasive cultural bias \cite{adilazuarda2024towards}. Researchers have found that LLMs often exhibit a preference for Western cultural entities and their opinions are more aligned with Western norms \cite{palta2023fork, ryan2024unintended}.



Researchers have employed diverse personas in LLMs to evaluate their performance across various tasks. Recent studies investigate how personas influence different aspects of model behavior \cite{de2024helpful, beck2024sensitivity}. Findings suggest that LLMs, when equipped with specific personas, can help reduce social biases \cite{kamruzzaman2024promptingtechniquesreducingsocial} and enhance zero-shot learning in subjective tasks \cite{beck2024sensitivity}. Conversely, other research indicates that personas can intensify the toxicity of model generations \cite{deshpande-etal-2023-toxicity} and task performance may vary based on the demographic attributes of the persona, such as gender and race \cite{salewski2024context}. This raises concerns that personas might not only improve performances but also perpetuate stereotypes. Previous studies have explored the effects of personas on various tasks, including sentiment analysis, hate speech detection sports understanding, MMLU, TruthfulQA, Bias Benchmark for QA, and ETHICS \cite{beck2024sensitivity, gupta2023bias, de2024helpful, mukherjee2024cultural}.

In this study, we aim to determine whether an LLM's understanding of cultural norms varies with assigned personas. It is evident from previous research that an LLM’s limited cultural knowledge can impact its predictions of cultural norms \cite{rao2024normad}. We investigate how the cultural knowledge that LLMs already possess might be influenced by the persona. To achieve this, we use two cultural norm datasets and assign 36 sociodemographic personas to four LLMs. \Cref{fig:example} illustrates how Llama 3's interpretation of social norms can change based on gender. 

The contributions of this paper are the following.
\begin{enumerate}

    \item
We present a comprehensive study examining how the interpretation of cultural norms by LLMs changes based on personas. In our research, we employed 36 distinct personas and four state-of-the-art LLMs across two social norm datasets.

\item 
Our study demonstrates that assigning personas leads to shifts in prediction accuracy, where socially preferred groups (e.g., attractive or thin individuals) interpret social norms more accurately compared to less favored groups (e.g., unattractive or fat individuals). 

\item 
We observe bias in the interpretation of cultural norms, where personas within a similar sociodemographic group can exhibit different cultural interpretations. For example, our study found that `a woman' persona interprets social norms more accurately than `a man' persona, as illustrated in \Cref{fig:example}. These findings suggest that although LLMs can tailor responses, their adaptability is influenced by inherent biases associated with these personas. 

    \item 
We also observe that the selection of personas is crucial; some personas lead to a better interpretation of norms, while others do not. For instance, in exploring cultural norms, personas that are culturally aware, such as a `travel expert' tend to perform better than others. 
    
\end{enumerate}

%% file: related_work.tex
\section{Related Work}

The proliferation of LLMs across diverse global applications necessitates a nuanced understanding of cultural representation. Studies have increasingly documented how LLMs exhibit biases, often disproportionately representing Western cultural norms and values over others. For instance, investigations into the cultural preferences of LLMs reveal a distinct bias towards Western cultural entities and etiquettes, aligning LLM outputs with Western societal norms while neglecting non-Western perspectives \cite{adilazuarda2024towards, liu2024culturally, ramezani2023knowledge, bhatt2024extrinsic}. 
Recent research has focused on enhancing the cultural competence of LLMs by integrating diverse cultural datasets into model training, thus fostering a more balanced cultural representation. For example, \citet{li2024culture} explores methods for incorporating a broader spectrum of cultural data, which helps in moderating the cultural biases inherent in LLMs. Probing techniques further facilitate an in-depth understanding of the cultural knowledge embedded in LLMs \cite{arora2022probing}. 
Moreover, the concept of cross-cultural alignment in LLMs emphasizes adjusting model outputs to reflect a fair and equitable representation of diverse cultures \cite{lee2024exploring, fung2024massively}. 
Central to these efforts are datasets specifically designed to evaluate and enhance the cultural adaptability of LLMs. Researchers proposed datasets that are designed to explore the ability of LLMs to handle culturally specific norms across country and continent levels. The NORMAD \cite{rao2024normad} and EtiCor \cite{dwivedi2023eticor} datasets, for instance, provide extensive resources for probing and benchmarking the cultural knowledge and sensitivities of LLMs across myriad cultural contexts.  




Assigning different sociodemographic personas to LLMs (\textit{sociodemographic prompting}) is a common approach for studying cultural biases \cite{beck2024sensitivity, mukherjee2024cultural}. Recent research has explored biases in persona-assigned LLMs. For instance, \citet{gupta2023bias} discovered that assigning a disabled persona leads to lower reasoning task scores. Additionally, personas can increase toxicity in LLMs \cite{deshpande-etal-2023-toxicity} and reinforce social stereotypes \cite{plaza2024angry}. Conversely, some studies have shown that specific personas can enhance model performance and reduce social biases \cite{beck2024sensitivity, kamruzzaman2024promptingtechniquesreducingsocial}. Further research has investigated the performance and sensitivity of LLMs to different personas, revealing that LLMs are responsive to these changes \cite{de2024helpful, kamruzzaman2024exploring, beck2024sensitivity}. These studies cover a range of tasks from sentiment analysis to reasoning and involve assigning various types of personas, from direct sociodemographic categories like man and woman to those conditioned by attributes like a favorite programming language or an imagined persona's house number.

%% file: dataset_creation.tex
\section{Datasets}
In this study, we use two cultural norm datasets: 1)~NORMAD \cite{rao2024normad} and 2)~EtiCor \cite{dwivedi2023eticor}. 

\paragraph{NORMAD dataset.} This dataset comprises 2.6k stories depicting social and cultural norms from 75 countries. The NORMAD dataset is created from cultural data in Cultural Atlas\footnote{https://culturalatlas.sbs.com.au} and uses an automated, human-in-the-loop process to generate narrative stories. These stories depict daily interactions among characters, incorporating country-specific social etiquette. They categorized these social etiquettes into four categories namely basic etiquette, eating, visiting, and gift-giving etiquette.
\paragraph{EtiCor dataset.} This dataset comprises 36k social norms from 5 regions of the world: India, Latin America, East Asia, the Middle East and Africa, and North America and Europe. 
This dataset mostly covers four major types of etiquette, namely dining and festivals, visits and social interactions, travel, and business. 

\begin{table*}[h]
    \centering
    {\small
    \begin{tabular}{>{\raggedright\arraybackslash}p{4cm}>{\raggedright\arraybackslash}p{10cm}}
        \hline
        \textbf{Persona Group} & \textbf{Personas} \\
        \hline
        Gender & a man, a woman, a transgender man, a transgender woman, a non-binary person \\ \hline
        Disability & a physically disabled person, an able-bodied person \\
        \hline
        Age & an old person, a young person \\
        \hline
        Physical Appearance & a fat person, a thin person \\
        \hline
        Race & a White person, a Black person \\
        \hline
        Skin Tone & a dark-skinned person, a light-skinned person  \\
        \hline
        Education Level &  a person who is a high school graduate, who has an associate degree in college, a person who has a bachelor’s degree in college, a person who has a doctoral degree, a person who has less than a high school degree\\
        \hline
        Profession &  a doctor, an engineer, a security guard, a cleaner\\
        \hline
        Beauty &  an attractive person, an unattractive person\\
        \hline
        Social Class &  a lower-class person, a middle-class person, a higher-class person \\
        \hline
        Income Level &  a low-income person, a high-income person \\
        \hline
        Cultural Awareness &  a culturally aware person, a well-traveled person, a homebound person, a globetrotter\\
        \hline
    \end{tabular}
    }
    \caption{The 36 Personas across 12 sociodemographic groups.}
    \label{tab:all_persona_list}
\end{table*}

%% file: experimental_setup.tex
\section{Experimental Setup}
\paragraph{Personas.} We use 36 personas across 12 sociodemographic groups listed in \Cref{tab:all_persona_list}. 
\paragraph{Models.} We use four different LLMs in our study namely GPT-4o-mini~\citep{achiam2023gpt}, using \texttt{gpt-4o-mini} checkpoint on the OpenAI API, Llama3-8B \citep{touvron2023llama} using the \texttt{meta-llama/Meta-Llama-3-8B-Instruct} checkpoint on Huggingface, Gemma2-27B \citep{team2024gemma} using the \texttt{google/gemma-2-27b-it} checkpoint on Huggingface, and Mistral-7B \citep{jiang2023mistral} using the \texttt{mistralai/Mistral-7B-Instruct-v0.3} checkpoint on Huggingface.  


\paragraph{Assigning personas and Prompting instructions.}
We assign personas to our LLMs using the three prompting templates listed in \Cref{tab:prompting_templates}. We adopted these three prompting templates from \citet{gupta2023bias}. In our experiments, each model was prompted 108 times (36 personas times 3 templates).  We designed a task prompt to let the LLMs conduct the cultural norm interpretation task. For each social norm in the NORMAD dataset, we append the following task prompt after the persona prompt: \textit{``
Read the story and answer if the action of the character in the story is socially acceptable in \{country\}. Answer with `yes', `no', or `neutral' and omit explanations.
Story: \{story\}''}

For the EtiCor dataset, we use the same persona templates but we instruct the model to answer with only `yes' and `no' as there is no `neutral' label in the EtiCor dataset. For the EtiCor dataset, we append the following task prompt after the persona prompt: \textit{``
Read the sentence and say whether this is a good etiquette or a socially acceptable norm in \{region\}. If the etiquette is socially acceptable in \{region\}, respond with `yes', if not acceptable respond with `no' and omit explanations.
Sentence: \{sentence\}''}. 
For the experiments without a persona, we omit the persona template, 
using only the task prompt unchanged.
Additionally, for these non-persona experiments, we use three prompting templates: one baseline and two that involve rephrased versions of the baseline, see \Cref{tab:without_persona_prompting} and \Cref{tab:without_persona_prompting_eticor} in \Cref{app:without_persona_template}.


\begin{table*}[h]
    \centering
    {\small
    \begin{tabular}{c|p{12cm}}
        \hline
        \textbf{Prompt Number} & \textbf{Persona Template} \\
        \hline
        P1 & You are \{persona\}. Your responses should closely mirror the knowledge and abilities of this persona. \\
        \hline
        P2 & Adopt the identity of \{persona\}. Answer the questions while staying in strict accordance with the nature of this identity. \\
        \hline
        P3 & Take the role of \{persona\}. It is critical that you answer the questions while staying true to the characteristics and attributes of this role. \\
        \hline
    \end{tabular}
    }
    \caption{We utilize the three distinct Persona Instructions from \citet{gupta2023bias} to assign persona (e.g., a fat person) to an LLM. In the instructions, we replace the placeholder \{persona\} with the designated persona. }
    \label{tab:prompting_templates}
\end{table*}

%% file: results.tex
\section{Results and Discussion}

\subsection{Cultural Norm Interpretation Sensitivity} We investigate the sensitivity of cultural norm predictions, specifically the extent to which LLMs' predictions vary when instructed to respond from viewpoints characterized by specific sociodemographic backgrounds.

\begin{table}[h]
\centering
{\small
\setlength{\tabcolsep}{3.5pt}
\begin{tabular}{|l|c|c|}
\hline
\multirow{2}{*}{\textbf{Model}} & \multicolumn{1}{c|}{\textbf{NORMAD}} & \multicolumn{1}{c|}{\textbf{EtiCor}} \\
\cline{2-3}
 & \textbf{Acc} & \textbf{Acc} \\
\hline
Llama3 with Persona & 46.06 & 59.23 \\ 
Llama3 without Persona & 45.75 & 54.00 \\ \hline
Gemma2 with Persona & 56.87 & 66.07 \\ 
Gemma2 without Persona & 57.50 & 55.00 \\ \hline
Mistral with Persona & 30.45 & 35.56 \\
Mistral without Persona & 16.52 & 12.46 \\ \hline
GPT-4o-mini with Persona & 55.74 & 72.13 \\
GPT-4o-mini without Persona & 58.03 & 73.64 \\ \hline
\end{tabular}
}
\caption{Comparison of model accuracies for NORMAD and EtiCor datasets, with (averaged across all personas) and without persona. All these results are averaged across all three prompting templates. }
\label{tab:average-model-comparison-without-persona}
\end{table}


\paragraph{Cultural norm interpretation changed when personas are used.}

\begin{table*}[h]
\centering
{\small
\setlength{\tabcolsep}{3.0pt}
\begin{tabular}{|l|c|c|c|c||c|c|c|c|}
\hline
\multirow{2}{*}{\textbf{Persona}} & \multicolumn{4}{c||}{\textbf{NORMAD Dataset}} & \multicolumn{4}{c|}{\textbf{EtiCor Dataset}} \\
\cline{2-9}
 & \textbf{Llama3} & \textbf{Gemma2} & \textbf{Mistral} & \textbf{GPT-4o-mini} & \textbf{Llama3} & \textbf{Gemma2} & \textbf{Mistral} & \textbf{GPT-4o-mini} \\
\hline
Young & 47.45 & 58.50 & \textcolor{orange}{\textbf{28.54}} & 57.65 & 60.54 & 67.66 & \textcolor{orange}{\textbf{32.73}} & 73.10 \\ \hline
Old & 46.89 & 58.30 & \textcolor{orange}{\textbf{21.55}} & 57.05 & 60.47 & 67.52 & \textcolor{orange}{\textbf{26.05}} & 73.55 \\ \hline \hline
Less than High School & 46.89 & 58.37 & 34.24 & 57.29 & 59.19 & \textcolor{orange}{\textbf{64.90}}& 39.07 & \textcolor{orange}{\textbf{71.86}} \\ \hline
High School Graduate & 47.30 & 57.17 & \textcolor{orange}{\textbf{38.22}} & 57.35 & 59.59 & 67.29 & \textcolor{orange}{\textbf{42.10}} & 73.21 \\ \hline
Associate Degree & 46.25 & 58.06 & \textcolor{orange}{\textbf{21.59}} & 57.79 & 60.60 & 66.60 & \textcolor{orange}{\textbf{28.93}} & 73.86 \\ \hline
Bachelor's Degree & 47.35 & 58.37 & 33.21 & 57.93 & 59.92 & 67.20 & 37.81 & 73.80 \\ \hline
Doctoral Degree & 46.63 & 58.24 & 28.83 & 58.11 & 60.77 & \textcolor{orange}{\textbf{67.79}} & 35.01 & \textcolor{orange}{\textbf{74.45}} \\ \hline \hline
Woman & \textcolor{orange}{\textbf{46.69}} & \textcolor{orange}{\textbf{58.13}} & \textcolor{orange}{\textbf{39.23}} & 56.69 & 58.75 & \textcolor{orange}{\textbf{66.59}} & \textcolor{orange}{\textbf{42.84}} & 72.99 \\ \hline
Man & \textcolor{orange}{\textbf{43.47}} & \textcolor{orange}{\textbf{53.82}} & \textcolor{orange}{\textbf{25.61}} & \textcolor{orange}{\textbf{57.68}} & 59.44 & 64.99 & \textcolor{orange}{\textbf{25.56}} & \textcolor{orange}{\textbf{73.25}} \\ \hline
Transgender Woman & 44.78 & 54.21 & \textcolor{orange}{\textbf{32.51}} & 51.30 & 58.77 & 64.50 & 30.98 & 71.29 \\ \hline
Transgender Man & 44.03 & 54.03 & \textcolor{orange}{\textbf{37.97}} & \textcolor{orange}{\textbf{51.01}} & 57.99 & 64.08 & 34.76 & \textcolor{orange}{\textbf{70.39}} \\ \hline
Non-binary & 44.35 & 54.23 & 41.46 & 54.17 & 56.39 & \textcolor{orange}{\textbf{63.49}} & 38.37 & 71.77 \\ \hline \hline
Cleaner & 46.93 & 56.77 & \textcolor{orange}{\textbf{22.84}} & 57.60 & 60.51 & 66.07 & \textcolor{orange}{\textbf{31.49}} & 73.32 \\ \hline
Doctor & 46.60 & 56.96 & 30.14 & 57.03 & 59.84 & 66.10 & 37.24 & 73.10 \\ \hline
Enginner & 46.97 & 56.86 & \textcolor{orange}{\textbf{41.57}} & 56.55 & 59.32 & 65.70 & \textcolor{orange}{\textbf{44.96}} & 72.02 \\ \hline
Security Guard & 47.21 & 56.94 & 36.27 & 56.05 & 58.45 & 65.16 & 41.51 & 72.27 \\ \hline \hline
Attractive & \textcolor{orange}{\textbf{45.46}} & \textcolor{orange}{\textbf{56.34}} & \textcolor{orange}{\textbf{17.39}} & 56.87 & 59.92 & 66.17 & 28.28 & 73.15 \\ \hline
Unattractive & \textcolor{orange}{\textbf{43.85}} & \textcolor{orange}{\textbf{53.64}}
 & \textcolor{orange}{\textbf{22.09}} & 55.44 & 59.53 & 65.15 & 28.17 & 73.39 \\ \hline \hline
Thin & 45.56 & \textcolor{orange}{\textbf{56.96}} & \textcolor{orange}{\textbf{25.14}} & \textcolor{orange}{\textbf{57.79}} & 59.44 & 65.71 & 33.98 & 73.72 \\ \hline
Fat & 45.45 & \textcolor{orange}{\textbf{54.06}} & \textcolor{orange}{\textbf{29.54}} & \textcolor{orange}{\textbf{54.63}} & 58.92 & 64.88 & 34.65 & 72.41 \\ \hline \hline
Light-skinned & 45.31 & 56.46 & \textcolor{orange}{\textbf{31.59}} & \textcolor{orange}{\textbf{57.79}} & \textcolor{orange}{\textbf{58.41}} & 65.24 & \textcolor{orange}{\textbf{37.99}} & 73.57 \\ \hline
Dark-skinned & 44.79 & 55.58 & \textcolor{orange}{\textbf{36.62}} & \textcolor{orange}{\textbf{55.92}} & \textcolor{orange}{\textbf{56.92}} & 64.64 & \textcolor{orange}{\textbf{42.71}} & 72.55 \\ \hline \hline
Able-bodied & 45.80 & 54.37 & \textcolor{orange}{\textbf{35.27}} & \textcolor{orange}{\textbf{56.30}} & \textcolor{orange}{\textbf{57.61}} & 64.19 & \textcolor{orange}{\textbf{38.39}} & 73.35 \\ \hline
Physically-disabled & 44.66 & 54.29 & \textcolor{orange}{\textbf{39.49}} & \textcolor{orange}{\textbf{50.70}} & \textcolor{orange}{\textbf{53.96}} & 63.49 & \textcolor{orange}{\textbf{42.16}} & 71.17 \\ \hline \hline
Black & 45.74 & 57.50 & \textcolor{orange}{\textbf{15.55}} & 57.17 & 61.45 & 67.08 & \textcolor{orange}{\textbf{25.45}} & 73.37 \\ \hline
White & 46.45 & 57.94 & \textcolor{orange}{\textbf{23.36}} & 57.63 & 61.08 & 66.77 & \textcolor{orange}{\textbf{31.17}} & 73.56 \\ \hline \hline
Lower-Class & 46.25 & 58.86 & 25.52 & 57.06 & 60.13 & 66.82 & 31.50 & 72.98 \\ \hline
Middle-Class & 45.92 & \textcolor{orange}{\textbf{56.53}} & \textcolor{orange}{\textbf{29.24}} & 57.88 & 59.46 & 66.26 & \textcolor{orange}{\textbf{35.86}} & 73.61 \\ \hline
Upper-Class & 46.84 & \textcolor{orange}{\textbf{59.06}} & \textcolor{orange}{\textbf{18.31}} & 58.69 & 60.76 & 66.95 & \textcolor{orange}{\textbf{24.73}} & 73.50 \\ \hline \hline
Low-Income & 46.64 & 58.94 & \textcolor{orange}{\textbf{31.51}} & 57.34 & \textcolor{orange}{\textbf{59.71}} & \textcolor{orange}{\textbf{66.37}} & \textcolor{orange}{\textbf{36.18}} & 73.12 \\ \hline
High-Income & 45.78 & 57.51 & \textcolor{orange}{\textbf{35.90}} & 57.97 & \textcolor{orange}{\textbf{55.76}} & \textcolor{orange}{\textbf{63.85}} & \textcolor{orange}{\textbf{40.07}} & 73.00 \\ \hline \hline
Culturally Aware & 46.74 & 58.37 & 36.00 & 58.10 & 59.29 & 66.01 & 40.50 & 74.24 \\ \hline
Travel Expert & 46.97 & 58.78 & 23.76 & 58.22 & \textcolor{orange}{\textbf{60.21}} & \textcolor{orange}{\textbf{68.65}} & 33.38 & \textcolor{orange}{\textbf{74.61}} \\ \hline
Well-Traveled & 46.34 & \textcolor{orange}{\textbf{59.99}} & 30.37 & \textcolor{orange}{\textbf{58.25}} & 60.07 & 68.46 & 38.65 & 74.61 \\ \hline
Homebound & 45.82 & 57.92 & 33.99 & 57.33 & 58.76 & 66.28 & 39.28 & 72.19 \\ \hline
Globetrotter & \textcolor{orange}{\textbf{47.01}} & 58.98 & \textcolor{orange}{\textbf{39.83}} & 58.24 & 59.48 & 67.96 & \textcolor{orange}{\textbf{44.78}} & 74.32 \\ \hline 
\end{tabular}
}
\caption{Comparison of model accuracy across 
personas and datasets 
averaged across all 
prompting templates. As the table is big we colored (orange) the results that we focused on in the body text of the paper.
}
\label{tab:combined-persona-model-comparison}
\end{table*}

In \Cref{tab:average-model-comparison-without-persona}, we present the accuracy results for both datasets with and without personas. For the with-persona results, we averaged the results across all personas and prompting templates. As shown in \Cref{tab:average-model-comparison-without-persona}, accuracy varies depending on the model and dataset. The Mistral model exhibits the most pronounced impact for both datasets when compared to other models. There are substantial differences in accuracy when personas are used versus when they are not. Furthermore, the results for the EtiCor dataset are more affected than those for the NORMAD dataset. We also notice that GPT-4o-mini is the least affected on average for both datasets.

\paragraph{Cultural norm interpretation differs within similar sociodemographic groups.} 

In \Cref{tab:combined-persona-model-comparison}, we present the results for each persona averaged across all the prompting templates. 
We notice differences in accuracy among similar sociodemographic profiles (e.g., man and woman). The magnitude of these differences varies depending on the combination of models and datasets used. Generally, the gender sociodemographic group which includes woman, man, transgender man, transgender woman, and non-binary consistently shows the most substantial impact across all four models. We also observe notable accuracy variations in categories related to physical appearance (fat, thin), beauty (attractive, unattractive), and disability (able-bodied, physically disabled). 
It appears that similar sociodemographic profiles tend to exhibit greater changes in accuracy in the NORMAD dataset than in the EtiCor dataset.

\paragraph{%
All regions are sensitive to sociodemographic prompting but no region is consistently more sensitive across both datasets and all models.
} 
Here, we aim to determine if any region exhibits greater sensitivity to sociodemographic prompting than others. In \Cref{tab:region-model-comparison-personas}, we present the results by region, both with and without the use of personas. The EtiCor dataset includes norms from five regions. Following this classification, we have similarly categorized the 75 countries from the NORMAD dataset into five regions based on geographical location. Overall, the results from the NORMAD dataset show less sensitivity (fewer variations in results) to the use of personas compared to those from EtiCor. We notice that the Mistral model is particularly sensitive to sociodemographic prompting.



\begin{table*}[h]
\centering
{\small
\setlength{\tabcolsep}{2.5pt}
\begin{tabular}{|l|c|c|c|c||c|c|c|c|}
\hline
\multirow{2}{*}{\textbf{Region}} & \multicolumn{4}{c||}{\textbf{NORMAD Dataset}} & \multicolumn{4}{c|}{\textbf{EtiCor Dataset}} \\
\cline{2-9}
 & \textbf{Llama3} & \textbf{Gemma2} & \textbf{Mistral} & \textbf{GPT-4o-mini} & \textbf{Llama3} & \textbf{Gemma2} & \textbf{Mistral} & \textbf{GPT-4o-mini} \\
\hline
East Asia (WP) & 48.76 & 62.13 & 30.11 & 61.52 & 58.21 & 63.58 & 42.00 & 75.15 \\ \hline
East Asia (W/OP) & 47.79 & 61.76 & 16.97 & 62.31 & 55.20 & 54.20 & 10.10 & 75.60 \\ \hline
India (WP) & 37.34 & 59.71 & 26.19 & 58.79 & 64.87 & 69.88 & 35.70 & 76.08 \\ \hline
India (W/OP) & 36.85 & 60.47 & 16.53 & 59.69 & 54.75 & 54.85 & 10.65 & 75.90 \\ \hline
Latin America (WP) & 46.51 & 50.46 & 32.80 & 52.48 & 52.81 & 60.53 & 34.98 & 68.97 \\ \hline
Latin America (W/OP) & 47.36 & 52.75 & 13.43 & 54.69 & 52.80 & 53.15 & 12.90 & 69.45 \\ \hline
Middle East and Africa (WP) & 42.19 & 56.71 & 28.96 & 53.91 & 56.74 & 63.38 & 37.82 & 72.03 \\ \hline
Middle East and Africa (W/OP) & 43.91 & 58.62 & 16.47 & 55.82 & 53.00 & 55.20 & 10.90 & 71.95 \\ \hline
North America-Europe (WP) & 48.60 & 55.13 & 32.14 & 56.90 & 63.78 & 73.40 & 27.32 & 75.02 \\ \hline
North America-Europe (W/OP) & 49.38 & 55.16 & 17.12 & 57.89 & 55.25 & 57.60 & 10.75 & 74.30 \\ \hline
\end{tabular}
}
\caption{Comparison of model accuracies across different regions for NORMAD and EtiCor datasets, where we present With Persona results as \textbf{WP} and Without Persona results as \textbf{W/OP}. All these results are averaged across all three prompting templates. }
\label{tab:region-model-comparison-personas}
\end{table*}

\subsection{Performance} 
Here, we investigate whether using a persona helps in the accurate interpretation of cultural norms. 


\paragraph{Performance improvement depends on dataset, model, and persona combinations.}



In the NORMAD dataset, the results are somewhat mixed. \Cref{tab:average-model-comparison-without-persona} shows that Llama3 and Mistral perform better with personas, whereas Gemma2 and GPT-4o-mini do not exhibit improved performance with personas, although the performance differences are slight. However, these results don't provide the full picture. Upon examining \Cref{tab:combined-persona-model-comparison}, it becomes clear that performance varies greatly depending on the personas. When cultural awareness is considered a factor of sociodemographic control, personas such as `well-traveled', `globetrotter', and `travel expert' consistently outperform others. Additionally, personas that are socially more desirable, such as `an attractive person', `a thin person', `an able-bodied person', and `an engineer', generally perform well. 
For the EtiCor dataset, most models show improved performance with personas, as indicated in \Cref{tab:average-model-comparison-without-persona}. Here too, the effectiveness depends on the persona. Personas that denote cultural awareness tend to perform well, and we observe that individuals with higher educational qualifications generally outperform other personas. So, here performance improvement really depends on which persona we use.




\paragraph{Model choice matters a lot. } Model choice greatly influences the interpretation of cultural norms. On average, GPT-4o-mini outperforms other models, while Mistral shows lesser accuracy for both datasets (refer to \Cref{tab:average-model-comparison-without-persona}). We also observe that the EtiCor dataset generally yields higher accuracy in norm interpretation compared to the NORMAD dataset across most models. In persona-specific comparisons (\Cref{tab:combined-persona-model-comparison}), performance varies across different models. For the NORMAD dataset, the highest recorded accuracy is 59.99\%, achieved by the Gemma2 model. Conversely, for the EtiCor dataset, GPT-4o-mini leads with a maximum accuracy of 74.61\%. Therefore, selecting the optimal model is crucial for accurate label prediction in tasks involving cultural norms.




\begin{figure}[t]
\centering
\includegraphics[width=1.0\linewidth]{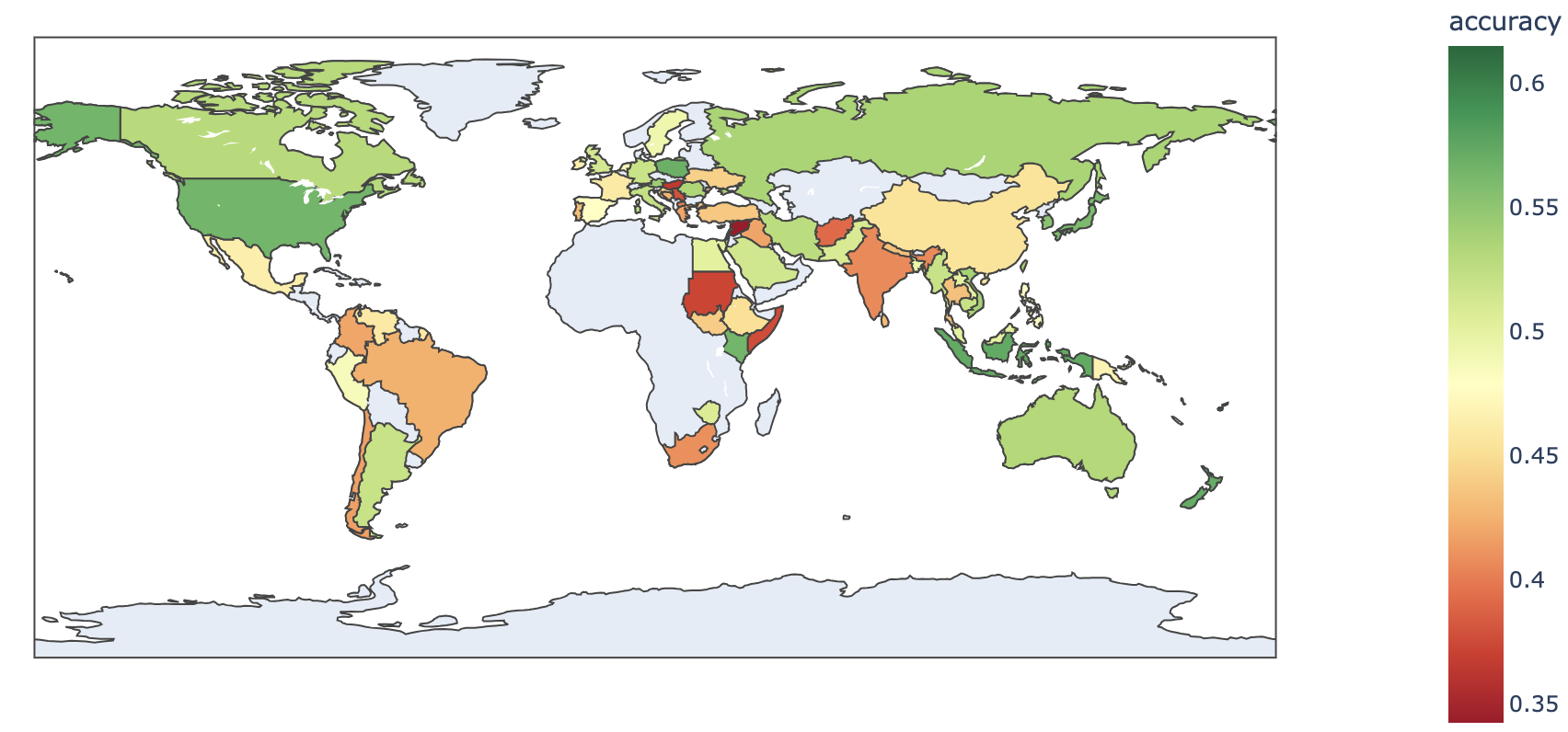}
\caption{County-level accuracy for NORMAD dataset averaged across all the models, personas, and prompting templates.}
\label{fig:world_map}
\end{figure}

\paragraph{Continent and Performance. } From \Cref{tab:region-model-comparison-personas}, it is evident that in the East Asia region, most models (with the exception of GPT-4o-mini) perform well with personas in both datasets. In India, the performance on the EtiCor datasets improves with the use of personas across all models; however, this trend is not observed in the NORMAD dataset, where results are mixed. The results for Latin America and the Middle East and Africa regions are somewhat noisy, with no clear patterns observed. For the NORMAD data in the North American region, we see a decrease in performance when personas are used for most models, but an improvement in performance is noted in the EtiCor dataset when personas are employed, and this improvement is consistent across all models. \Cref{fig:world_map} depicts the country-level results for the NORMAD dataset, showing no distinct pattern that indicates one region's countries performed better than others; rather, the results are generally mixed.



\subsection{Bias in Cultural Norm Interpretation} We observe variations in prediction sensitivity across different sociodemographic prompts. Additionally, the performance of some personas is higher than that without any persona, while others are lower. We will examine whether prediction rates change between similar sociodemographic groups (e.g., able-bodied persons versus physically disabled persons) is due to biases in LLMs. As shown in \Cref{tab:combined-persona-model-comparison}, there are noticeable differences in the prediction rates of similar sociodemographic groups. We have measured correlations and statistical significance using Kendall's $\tau$ test~\cite{kendall-1938-new} for a few selected samples where we observe large differences (as in \Cref{tab:combined-persona-model-comparison}), presented in \Cref{tab:statistical_test}. 

\paragraph{NORMAD dataset. } 


For the `young' and `old' personas, all models tend to perform well for the `young' persona. This indicates an ageism bias, where decisions are based more on age than on quality. For the Mistral model, we notice high differences in accuracies, indicating bias which is also statistically significant (see \Cref{tab:statistical_test}).

For the gender sociodemographic group, we notice prediction changes across all models. In all models except GPT-4o-mini, the prediction rate for the `woman' persona is higher than for the `man' persona, indicating a deeply embedded gender bias in LLMs. In the case of GPT-4o-mini, the prediction rates for `transgender woman' and `transgender man' personas are greatly lower than those for `woman' and `man' personas. When conducting a country-level analysis, we find that the `transgender woman' and `transgender man' personas perform poorly in interpreting cultural norms in Muslim-majority Arab countries such as Saudi Arabia, Iraq, and Iran. This reflects the general perception that transgender individuals are viewed negatively in these countries due to religious beliefs, a result that is also statistically significant.

\begin{table*}[h]
\centering
{\small
\setlength{\tabcolsep}{3.0pt}
\begin{tabular}{|l|l|l|c|c|l|}
\hline
\textbf{Group 1} & \textbf{Group 2} & \textbf{Model} & \textbf{$\tau$} & \textbf{$p$ } & \textbf{Dataset} \\ \hline
Young            & Old              & Mistral        &  0.210         &      \textbf{<0.001}     & NORMAD  \\ \hline
Woman            & Man              & Gemma2         &  0.112         &      \textbf{<0.001}     & NORMAD  \\ \hline
Man              & Transgender Man  & GPT-4o-mini    &  0.103         &      \textbf{<0.001}     & NORMAD  \\ \hline
Attractive       & Unattractive     & Llama3         &  0.045         &      0.104     & NORMAD  \\ \hline
Man              & Transgender Man  & GPT-4o-mini    &  0.092         &       \textbf{<0.001}    & EtiCor  \\ \hline
Low income       & High income      & Llama3         &  0.069         &     \textbf{<0.001}      & EtiCor  \\ \hline
White            & Black            & Mistral        &  0.071          &        \textbf{<0.001}   & NORMAD  \\ \hline
Attractive        & Unattractive & Gemma2     &   0.062        &      \textbf{<0.001}     & NORMAD  \\ \hline
Doctorate        & Less than High School & Gemma2     &    0.084       &      \textbf{<0.001}     & EtiCor  \\ \hline
\end{tabular}
}
\caption{Kendall's $\tau$ test results where we try to see if group 1 more accurately predicts the gold label than group 2. We use a significance level of $\alpha < 0.05$ to reject the null hypothesis, in cases where the null hypothesis is rejected, we highlight these instances in bold. }
\label{tab:statistical_test}
\end{table*}


Regarding the `attractive' and `unattractive' personas, we observe a consistent trend where the prediction rate for the `attractive' persona is higher than that for the `unattractive' persona across all models, except Mistral. This suggests an underlying beauty bias in models where decisions may be influenced by perceived attractiveness. However, as shown in \Cref{tab:statistical_test}, these results are not statistically significant for Llama3. A similar trend is evident with the `thin' and `fat' personas, where the `thin' persona's prediction rate is higher for all models except Mistral. This behavior highlights a bias in models that may make assumptions about individuals' capabilities based on their body size or physical appearance. We also see similar biased results for the skin tone group.



For the `able-bodied' and `physically disabled' personas, the prediction rates are higher for the `able-bodied' persona across all models, except for Mistral. This consistent pattern suggests an ableism or ability bias, indicating that the models perceive able-bodied individuals as superior or more capable in certain aspects compared to physically disabled individuals. Upon conducting a country-level analysis, it becomes apparent that the GPT-4o-mini model exhibits poorer performance for South Asian countries, such as Bangladesh, Nepal, and Pakistan, when using prompts associated with a physically disabled persona.



We also observe racial bias in sociodemographic prompting. When comparing the results of `Black' and `White' personas, it becomes evident that across all models, the `White' persona consistently outperforms the `Black' persona, indicating a racial bias in LLMs. This effect is particularly pronounced in African countries like Sudan, Somalia, and Kenya when analyzing country-level data for the Mistral model.

Similarly, in the sociodemographic group based on social class, the `upper-class' persona generally performs better than the `lower-class' persona, although the magnitude of this difference is not substantial. This pattern highlights the presence of bias, reflecting prejudgments or discriminatory attitudes based on a person's social class.

\paragraph{EtiCor dataset.} 

Similar to the observations in the NORMAD dataset, various social biases are evident in LLMs within the EtiCor dataset. Specifically, within the educational level sociodemographic group, personas holding doctoral degrees exhibit significantly higher prediction accuracy compared to those with less than a high school education across all models, except Mistral. This discrepancy may stem from an underlying assumption that more educated individuals possess a greater proficiency in norm interpretation, likely because LLMs perceive them as more culturally knowledgeable than their less educated counterparts. In region-level analyses, GPT-4o-mini and Gemma2 demonstrate lower performance for personas from the Middle East and Africa with a `less than high school' educational background. Additionally, an interesting trend emerges with the `low-income' and `high-income' personas, where most models tend to yield higher accuracy for the `low-income' persona.





\begin{figure*}[t]
\centering
\includegraphics[width=1.0\linewidth]{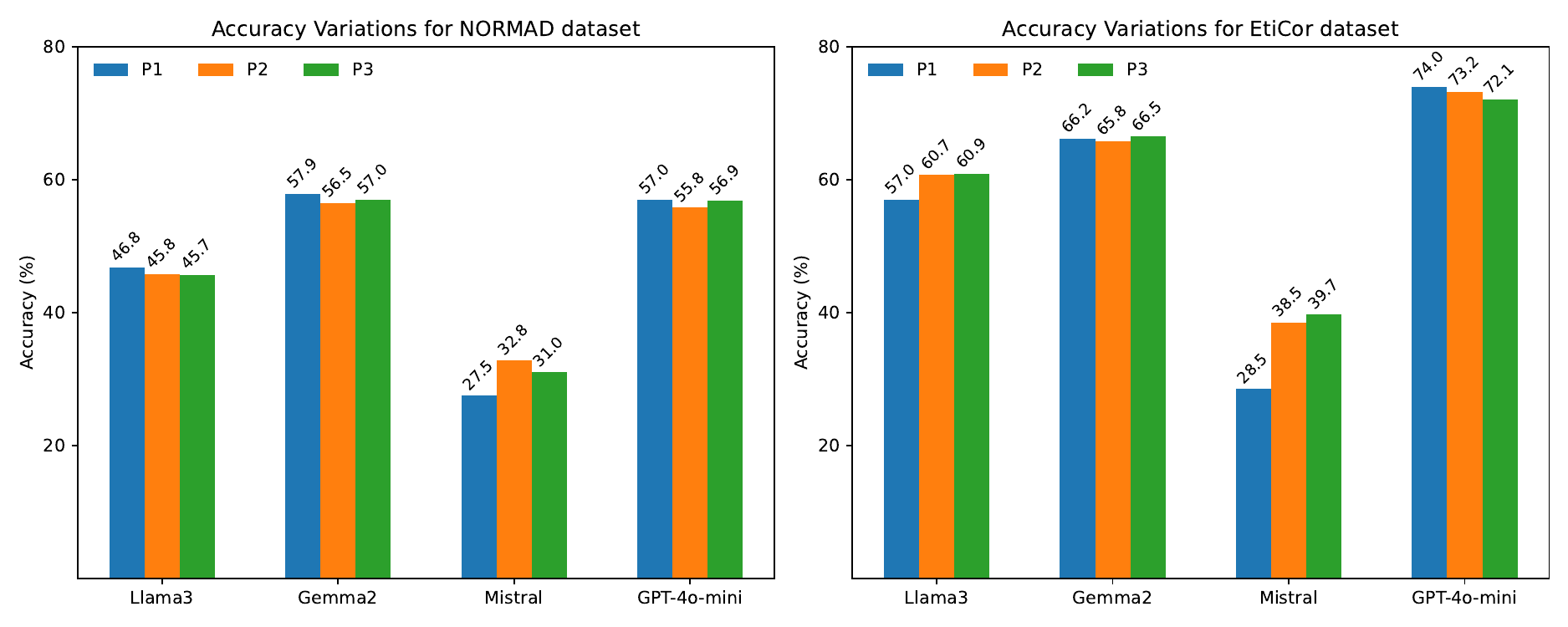}
\caption{Accuracy variations for all the three prompting templates, averaged across all the personas for each model.}
\label{fig:example_combibed_prompting}
\end{figure*}



\subsection{Robustness} We investigate how different prompting templates affect the prediction rates. We present the accuracy variation results averaged across all the personas in \Cref{fig:example_combibed_prompting}. 

\paragraph{All the models except Mistral look robust across the different prompting templates.} 
The accuracy differences among Llama3, Gemma2, and GPT-4o-mini are minor and remain consistent across most prompting templates. However, for the EtiCor dataset using the Llama3 model, we observe larger differences in accuracies between prompting 1 and promptings 2 and 3. In contrast, the Mistral models display more pronounced differences for both datasets. 
Our experimental setting shows better LLM robustness across sociodemographic prompting variations than what has been reported in past experiments~\citep{beck2024sensitivity}. 
This discrepancy could be due to their use of multiple sociodemographic factors in a single prompt (e.g., a person of gender `\{gender\}', race `\{race\}', age `\{age\}', education level `\{education\}'), whereas we employ only one sociodemographic profile at a time. Additionally, the choice of models could play a role, as they compared models of varying sizes, and smaller models tend to be less robust. 
InstructGPT, the largest model they used, showed greater robustness. We incorporated more recent model versions in our experiments, which may also account for the differences observed.



\subsection{Refusal rate change across personas}
Although we instructed the model to respond with `yes', `no', and `neutral' (for NORMAD only), models sometimes refuse to answer by expressing statements like \textit{`I'm sorry, but I cannot answer your question....', `I can't answer that. I don't know much about fancy stuff like traditions...', 
` As a transgender man, I don't have personal opinions or knowledge about ...', ` I'm an ai and don't have personal experiences...'}.
We see more refusals in the EtiCor dataset than in the NORMAD dataset as the EtiCor dataset has more data. Surprisingly we did not see any refusals for the GPT-4o-mini. We use regex patterns following \citet{de2024helpful} to extract refusal sentences, where we search for specific keywords or phrases (e.g., `I'm sorry', `ai', `sorry', `can't', `cannot', `don't', `do not', etc.) from the responses .


In our experiments, we observe that the Gemma2 model exhibits the highest refusal rate within the Eticor dataset, particularly for the `transgender man' persona (see \Cref{tab:invalid} in \Cref{app:refusal}). This trend indicates that refusals vary based on the persona involved. Furthermore, our analysis reveals that models are more likely to refuse to answer when the sociodemographic persona is perceived as less socially desirable. Specifically, for personas identified as `Black', `White', and `Dark-skinned', the model often declines to respond, citing concerns such as, ``I'm sorry, but I cannot complete your request as it relies on harmful stereotypes about race and cultural understanding''.

%% file: conclusion.tex
\section{Conclusion}

This study highlights the influence of persona assignment on cultural norm interpretation in LLMs. We found that LLMs exhibit varying accuracy based on sociodemographic personas, with socially desirable personas (e.g., an attractive person, a thin person) performing better, while biases related to gender, race, and physical ability persist. Our results show that some models are more sensitive to persona changes, and cultural norm interpretation is inconsistent even within similar sociodemographic groups. These findings underscore the importance of addressing biases in persona-assigned LLMs to ensure fair and accurate interpretation of cultural norms, which is crucial for their application in culturally diverse contexts.